\documentclass[conference]{IEEEtran}
\IEEEoverridecommandlockouts

\usepackage{tikz}
\usepackage{textcomp}
\usepackage{hyperref}
\usepackage{lipsum}

\usepackage{cite}
\usepackage{amsmath,amssymb,amsfonts}
\usepackage{algorithmic}
\usepackage{graphicx}
\usepackage{textcomp}
\usepackage{xcolor}
\def\BibTeX{{\rm B\kern-.05em{\sc i\kern-.025em b}\kern-.08em
    T\kern-.1667em\lower.7ex\hbox{E}\kern-.125emX}}
\begin{document}

\title{Evaluation of Deep Learning Models for Hostility Detection in Hindi Text
}

\author{
\IEEEauthorblockN{Ramchandra Joshi$^1$, Rushabh Karnavat$^1$, Kaustubh Jirapure$^1$, Raviraj Joshi$^2$}
\IEEEauthorblockA{
\textit{$^1$Pune Institute of Computer Technology, Pune, Maharashtra}\\
\textit{$^2$Indian Institute of Technology Madras, Chennai, Tamilnadu}\\
\{rbjoshi1309, rpkarnavat, kaustubhmjirapure, ravirajoshi\}@gmail.com}
}

\IEEEoverridecommandlockouts

\newcommand\copyrighttext{%
\footnotesize This work has been submitted to the IEEE for possible publication. Copyright may be transferred without notice, after which this version may no longer be accessible.}

\newcommand\copyrightnotice{%
\begin{tikzpicture}[remember picture,overlay]
\node[anchor=south,yshift=10pt] at (current page.south) {\fbox{\parbox{\dimexpr\textwidth-\fboxsep-\fboxrule\relax}{\copyrighttext}}};
\end{tikzpicture}%
}

\maketitle
\copyrightnotice
\IEEEpubidadjcol

\begin{abstract}
The social media platform is a convenient medium to express personal thoughts and share useful information. It is fast, concise, and has the ability to reach millions. It is an effective place to archive thoughts, share artistic content, receive feedback, promote products, etc. Despite having numerous advantages these platforms have given a boost to hostile posts. Hate speech and derogatory remarks are being posted for personal satisfaction or political gain. The hostile posts can have a bullying effect rendering the entire platform experience hostile. Therefore detection of hostile posts is important to maintain social media hygiene. The problem is more pronounced languages like Hindi which are low in resources. In this work, we present approaches for hostile text detection in the Hindi language. The proposed approaches are evaluated on the Constraint@AAAI 2021 Hindi hostility detection dataset. The dataset consists of hostile and non-hostile texts collected from social media platforms. The hostile posts are further segregated into overlapping classes of fake, offensive, hate, and defamation. We evaluate a host of deep learning approaches based on CNN, LSTM, and BERT for this multi-label classification problem. The pre-trained Hindi fast text word embeddings by IndicNLP and Facebook are used in conjunction with CNN and LSTM models. Two variations of pre-trained multilingual transformer language models mBERT and IndicBERT are used. We show that the performance of BERT based models is best. Moreover, CNN and LSTM models also perform competitively with BERT based models.

\end{abstract}


\section{Introduction}

Social media platforms generate a huge amount of content on day to day basis. The platforms are extensively used by artists to share their art and general people to share some aspects of their day to day life. It is used by people to voice their opinion or pass on useful information. 
However negative aspects of these social platforms are being exploited for personal gains. Extreme content and hate posts are taking over fun and information. The hostile posts are intentionally created to target a person, race, ethnicity, gender, or even a country. This is partly because the fear of exposing personal identity is no longer an issue. People create a fake or anonymous account to share offensive content. Such content and the trolling culture has resulted in mental trauma and violent clashes between divergent groups. 
The detection and removal of hostile content from internet platforms are of paramount importance.

The most simple approach for detecting hostile post is manual verification. However, it requires extensive manpower and is prone to subjective bias. A seemingly funny post for one gender may be offensive to another. Therefore automated technique for hostility detection is the way forward \cite{macavaney2019hate}. Initial approaches to automatic detection were based on the detection of hostile words \cite{warner2012detecting}. But it is difficult to maintain an exhaustive list of derogatory or offensive words. The context in which these words appear is also important for a text to be categorized as hostile. Recently different machine learning approaches have been proposed for hate or offensive post-detection \cite{malmasi2017detecting,zhang2018detecting}. 

In this work, we are concerned with hostile post-detection for the Hindi language. Hindi is the third most popular language in the world and the official language of India. It uses the Devanagari script in written form. The availability of Hindi keyboards on mobile and desktop applications has increased the usage of Hindi on social internet platforms. The amount of research in Hindi hostility detection is limited due to its low resource nature. 

In this paper, we describe the approaches for Hindi hostile post-detection on Constraint@AAAI 2021 dataset \cite{bhardwaj2020hostility}. This dataset consists of hostile posts collected from Twitter and Facebook in Hindi Devanagari script. This task is a multi-label multi-class classification problem where each post belongs to one or more subclass of hostile classes. The hostile labels in the shared dataset include fake, offensive, hate, and defamation. There is considerable overlap between these labels as the same post can belong to multiple categories. We formulate the problem as a multi-label text classification problem \cite{schmidt2017survey}. A hierarchical approach is used, we first distinguish between hostile and non-hostile posts. The post categorized as hostile is subsequently passed through individual hostile models for multi-label assignment. We do not rely on any external features and just use the textual content of the post for classification. We evaluate different deep learning algorithms for text classification based on Convolutional Neural Network (CNN), Long Short Term Memory (LSTM), and Bidirectional Encoder Representations from Transformers (BERT) \cite{devlin2018bert}. The models used are simple 1D CNN, multi-channel CNN, bi-directional LSTM, CNN+LSTM, mBERT, and IndicBERT. We also evaluate Hindi fast text word embeddings by Facebook \cite{joulin2016bag} and IndicNLP \cite{kunchukuttan2020ai4bharat} in combination with CNN and LSTM models.
The main contributions of this work are
\begin{itemize}
    \item We present a comparative analysis of BERT, CNN, and LSTM based models for the task of Hindi Hostility detection. Publicly available multilingual pre-trained language models mBERT and IndicBERT are used. We show that shallow neural networks based on CNN and LSTM perfom on par with deep BERT based models.
    \item Publicly available variants of Hindi fast text word embeddings are compared for their effectiveness in conjunction with CNN and LSTM based models.
\end{itemize}


\section{Related Work}

The amount of research for hostility detection in the Hindi language is considerably less as compared to English. 
A deep learning approach based on detecting offensive language in tweets was proposed in \cite{pitsilis2018detecting}. The content of the tweets along with some user-specific features was provided to multiple LSTM classifiers in an ensemble setting. 
Similar supervised learning approaches based on LSTM and SVM were also proposed in \cite{del2017hate}. They utilized features based on POS tags, word embeddings, sentiment polarity, and other lexical features to classify hate speech textual content on Facebook. 
The importance of using distributed representations in detecting hate speech was also established in \cite{djuric2015hate}.

A comparison of different models and input representations was performed in \cite{badjatiya2017deep}. The classifiers like logistic regression, random forest, SVMs, Gradient Boosted Decision Trees (GBDTs), and some neural networks like CNN and LSTM were used for hate speech detection task. The deep learning approaches were shown to perform significantly better than traditional models. The input representations used in conjunction with traditional models were char n-grams, TF-IDF vectors, and Bag of Words vectors (BoWV). The random word embeddings and GloVe embeddings fine-tuned during training were used as input to neural network based models.
Other deep learning based approaches for similar tasks were proposed in \cite{agrawal2018deep,cao2020deephate}. 
Such works for the Hindi language has been quite limited. Recent work \cite{jha2020dhot} presented an annotated dataset for offensive text in Hindi. A fast text-based classifier model was used for distinguishing between offensive and non-offensive tweets.  
Other datasets and approaches for offensive or hate speech detection in code mixed Hindi-English text was proposed in \cite{bohra2018dataset,mathur2018detecting}. 

Since our work is concerned with Hindi text classification we also review some of the works in this area. A comparison of different deep learning algorithms for Hindi text classification was presented in \cite{joshi2019deep}. Various deep learning models based on CNN, LSTM in combination with fast text word embeddings were evaluated. 
Resources for Indic languages in the form of datasets and models were presented in \cite{kunchukuttan2020ai4bharat}. Hindi was one of the languages for which monolingual corpus, fast text word embedding, and BERT based pre-trained models were publicly made available. We make use of the IndicNLP Hindi fast text word embedding model and the IndicBERT model introduced in this work.

\section{Dataset Details}

We have used the hostility detection dataset in the Hindi language made available under Constraint@AAAI 2021 Shared Task \cite{patwa2021overview}. The shared task mainly focused on the hostility detection on three major points, i.e. low-resource regional languages, detection in emergency situations, and early detection.

The dataset contains 8192 online posts collected from various social media platforms like Twitter, Facebook, WhatsApp, etc., and are tagged as hostile or non-hostile posts. Furthermore, hostile posts are classified as fake, hate, defamation, and offensive. 

Since each hostile post in the data can have more than one hostility dimension, so each post has a multi-label tag e.g, a sample sentence from train set - '\textit{Chandrayan mission success hone hi wala tha ki Modiji ne waha jakar disturb kar diya jiske karan scientist ki mahino ki mehnat bekar ho gayi thi}' is classified as hate, defamation and offensive. Though some of the hostile dimensions like hate and offensive may sound similar at the abstract-level, they totally have a different meaning. We define each hostile dimension in Table \ref{sent_label_info_table}, exact details about the dataset and labels is described in \cite{bhardwaj2020hostility}. 


The dataset statistics is described in Table \ref{data_sample_distribution_table}. Out of total 8192 online posts, 4358 examples are classified as non-hostile posts, while the remaining 3834 hostile posts are tagged with one or more hostile dimensions. In all the 3834 hostile posts, there are 1638 fake, 1132 hate, 1071 offensive, and 810 defamation posts and each post can belong to multiple hostile dimensions. Also, the organizers of the shared task have split the data into a pre-defined split of 70\% train, 10\% validation, and 20\% test set to ensure a uniform label distribution of the labels, and the same defined sets were made available to the participants.


\begin{table}
\centering
\caption{Sentence label information}
\label{sent_label_info_table}
\begin{tabular}{|c|c|}
\hline
Category & Description \\ \hline
Fake & A post or information which is false \\
Hate & A post spreading hate speech or encouraging violence. \\
Offensive & A post found to be insulting an individual or group \\
Defamation & A post lowering or degrading someone's reputation \\
Non-hostile & A post with no hostility. \\ \hline
\end{tabular}
\end{table}

\begin{table}
\centering
\caption{Dataset statistics and Label distribution. Columns Fake, hate, defame, and offense reflect the number of respective posts including multi-label cases. *Total denotes the total hostile posts.}
\label{data_sample_distribution_table}
\begin{tabular}{|c|ccccc|c|}
\hline
 &  &  & Hostile &  &  & Non-Hostile \\ \hline
 & Fake & Hate & Offense & Defame & *Total &  \\ \hline
Train & 1144 & 792 & 742 & 564 & 2678 & 3050 \\ \hline
Validation & 160 & 103 & 110 & 77 & 376 & 435 \\ \hline
Test & 334 & 237 & 219 & 169 & 780 & 873 \\ \hline
Overall & 1638 & 1132 & 1071 & 810 & 3834 & 4358 \\ \hline
\end{tabular}
\end{table}

\begin{table*}
\centering
\caption{Results on validation data showing overall F1-score and scores for individual labels.}
\label{val_result_table}
\begin{tabular}{cccccccc}
\hline
Model & Embedding & \multicolumn{1}{p{1.2cm}}{\centering Coarse \\Grained} & Defame & \multicolumn{1}{p{1cm}}{\centering Fake} & \multicolumn{1}{p{1cm}}{\centering Hate} & Offensive & \multicolumn{1}{p{1.2cm}}{\centering Fine \\Grained} \\ \hline
 & random & 95.81 & 33.33 & 72.15 & 47.01 & 55.60 & 55.58 \\
CNN & IndicNLP FastText & 94.45 & 28.91 & 73.80 & 40.21 & 50.65 & 52.61 \\
 & FBFastText & 96.55 & 25.19 & 73.68 & 46.08 & 55.02 & 54.36 \\ \hline
 & random & 95.44 & 33.61 & 73.15 & 50.18 & 54.63 & 56.53 \\
Multi-CNN & IndicNLP FastText & 96.18 & 33.09 & 70.80 & 46.01 & 53.12 & 54.24 \\
 & FBFastText & 96.42 & 28.76 & 68.67 & 42.15 & 52.58 & 51.68 \\ \hline 
 & random & 95.80 & 40.47 & 67.27 & 44.13 & 48.48 & 52.73 \\
BiLSTM & IndicNLP FastText & 93.96 & 30.65 & 69.78 & 41.66 & 49.23 & 51.51 \\
 & FBFastText & 94.95 & 25.60 & 68.68 & 45.53 & 50.22 & 51.42 \\ \hline
 & random & 95.06 & 44.56 & 68.42 & 44.87 & 52.25 & 54.88 \\
CNN+BiLSTM & IndicNLP FastText & 95.43 & 34.73 & 66.00 & 37.83 & 47.16 & 49.45 \\
 & FBFastText & 95.68 & 35.42 & 66.87 & 44.62 & 49.55 & 52.09 \\ \hline
IndicBert & - & 96.91 & 47.66 & 83.38 & 54.30 & 57.81 & 64.31 \\ \hline
mBert & - & 96.79 & 38.50 & 77.74 & 41.17 & 53.08 & 56.44 \\ \hline
Baseline & - & 84.11 & 43.57 & 68.15 & 47.49 & 41.98 & - \\ \hline
\end{tabular}
\end{table*}



\begin{figure}[htbp]
\centerline{\includegraphics[scale=0.5]{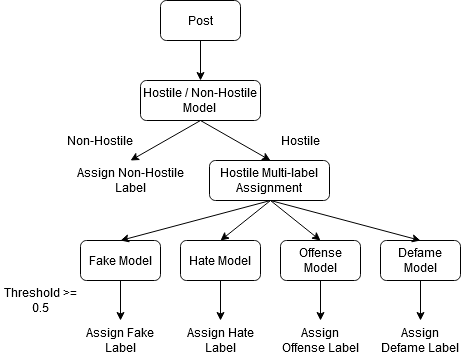}}
\caption{Flow of Multi-Class Multi-Label Classification}
\label{fig:multi_label_classification}
\end{figure}

\section{Model Architectures}

Traditionally, LSTM based models were preferred for text-based NLP tasks. However, recently Transformers based models have been shown to produce a state of the art results on most of the tasks \cite{vaswani2017attention}. In this work, we have explored a variety of deep learning models with variations in word embeddings for the Hindi Hostility detection task. Since each hostile post in the data has more than one hostility category label, the task is formulated as a multi-label classification task and individual models are trained for each label. Most importantly, while training all our models, we have used the one vs all strategy. The individual hostile class data were oversampled to balance the data during training. The non-hostile examples are also included in the all category for non-BERT experiments. We found degradation in performance after including non-hostile examples in the negative class for individual hostile BERT models. For the coarse-grained hostile and non-hostile classification, no oversampling was performed and all the hostile samples irrespective of their labels were taken together. The Adam optimizer and binary cross-entropy loss function are common across all the models. We employ minimal hyper-parameter tuning on the validation set. The validation loss is used to pick the best epoch.

\begin{itemize}
  \item \textbf{CNN :} This is a simple 1D CNN model processing sequence of word embeddings. The dimension of word embeddings is 300. The embedding layer is followed by a 1D convolutional layer using a kernel size of 4 and a filter size of 128. Next, a global max-pooling layer, a dropout layer with 0.3 drop rate, a dense layer of size 64, and a final dense layer of size 1 are used one after the another. The activation functions used with conv layer is relu and with dense layers is sigmoid. This same layer-activation pair is used in all the models.
  \item \textbf{Multi-CNN :} This is similar to the previous model but instead of using a single convolutional layer, 3 parallel convolutional layers are used. The kernel sizes for each of these are 2, 3, and 4 respectively. Other hyperparameters remain the same as the simple CNN model. The output of these convolutions is concatenated and passed to two dense layers of size 64 and 1 sequentially. 
  \item \textbf{Bi-LSTM :} This model is based on standard Bi-LSTM layers. The word embeddings are processed using a single Bi-LSTM layer with 128 units. The output hidden embedding is max-pooled over time and passed through dense layers of sizes 64 and 1. A recurrent dropout of 0.3 is also used.
  \item \textbf{CNN+Bi-LSTM :} The simple CNN and Bi-LSTM models described above are placed sequentially one after the other. The CNN processes the input word embeddings and other representations are passed to a Bi-LSTM layer. Subsequently, we have a global max-pooling layer followed by dense layers of sizes 64 and 1.
  \item \textbf{mBERT and IndicBERT}: The BERT is a bidirectional transformer language model trained on masked token prediction and next sentence prediction task. Since the language under consideration is Hindi we make use of the original multilingual BERT (mBERT) trained on 104 languages released by Google and IndicBERT trained on 10 Indian languages released by AI4Bharat. Both of these models are fine-tuned on the target task. The [CLS] token embedding is used for classification; it is followed by a single dense layer. 
\end{itemize}

\begin{table*}
\centering
\caption{Results on test data showing overall F1-score and scores for individual labels.}
\label{test_result_table}
\begin{tabular}{cccccccc}
\hline
Model & Embedding & \multicolumn{1}{p{1.2cm}}{\centering Coarse \\Grained} & Defame & \multicolumn{1}{p{1cm}}{\centering Fake} & \multicolumn{1}{p{1cm}}{\centering Hate} & Offensive & \multicolumn{1}{p{1.2cm}}{\centering Fine \\Grained} \\ \hline
 & random & 96.30 & 30.95 & 71.91 & 49.89 & 57.84 & 56.05 \\
CNN & IndicNLP FastText & 95.94 & 40.77 & 76.31 & 43.85 & 59.70 & 58.27 \\
 & FBFastText & 96.43 & 32.90 & 74.35 & 47.64 & 59.25 & 57.02 \\ \hline
 & random & 96.30 & 32.84 & 73.11 & 50.27 & 59.47 & 57.28 \\
Multi-CNN & IndicNLP FastText & \textbf{96.67} & 32.46 & 72.42 & \textbf{52.00} & 58.80 & 57.24 \\ 
 & FBFastText & 96.37 & 32.77 & 71.63 & 41.58 & 56.06 & 53.84 \\ \hline 
 & random & 96.18 & 36.08 & 71.89 & 49.88 & 54.80 & 56.26 \\
BiLSTM & IndicNLP FastText & 93.58 & 34.92 & 72.61 & 42.10 & 56.98 & 54.90 \\
 & FBFastText & 95.76 & 29.81 & 71.73 & 45.08 & 55.67 & 54.12 \\ \hline
 & random & 94.24 & 37.01 & 70.16 & 46.38 & 56.27 & 55.30 \\
CNN+BiLSTM & IndicNLP FastText & 96.18 & 41.14 & 70.39 & 46.49 & 56.59 & 56.21 \\
 & FBFastText & 95.64 & 35.40 & 70.67 & 50.19 & 54.50 & 55.72 \\ \hline
IndicBert & - & 96.06 & 40.35 & \textbf{79.93} & 51.61 & 59.38 & \textbf{61.29} \\ \hline
mBert & - & 96.18 & \textbf{45.30} & 75.77 & 46.66 & \textbf{64.13} & 60.59 \\ \hline
Baseline & - & 84.22 & 39.92 & 68.69 & 49.26 & 41.98 & 54.2 \\ \hline
\end{tabular}
\end{table*}

\section{Results and Discussion}

In this work, the performance of different deep learning models based on CNN, LSTM, and BERT was evaluated on the Hindi Hostility detection dataset. The input representation to basic models is distributed word vectors. We compare random initialization of word vectors and pre-trained fast text embeddings trained on Hindi corpus. The fast text word embeddings are based on subword units and well suited for noisy datasets. We fine-tune the pre-trained fast text word embeddings to adapt the model to the target corpus as the target domain is different from the domain on which the word vectors were trained. We use two variants of Hindi fast text embeddings released by Facebook and IndicNLP. The validation results in Table \ref{val_result_table} and test results in Table \ref{test_result_table} refer to the random initialization of word vectors as random, the fast text embedding released by IndicNLP as IndicNLP FastText and that released by Facebook as FBFastText. Weighted F1-score is used as a metric to compare the models. The execution of models is hierarchical in nature, refer Figure \ref{fig:multi_label_classification}. A given test sample is first categorized as hostile and non-hostile using the corresponding model. The hostile posts are then subsequently evaluated using individual hostile models for multi-label assignment.

The results in Table \ref{test_result_table} show that BERT based models give the best fine-grained F1-score whereas all the models give equally good coarse-grained scores. The IndicBERT performs slightly better than mBERT but there is no clear winner. The difference can also be attributed to minimal hyper-parameter tuning. The multi-CNN based models with any variation of word embedding have a slight edge over the other basic models. We report an absolute improvement of 11.84 \% in coarse-grained F1-scores and 7.09 \% in fine-grained F1-scores using IndicBERT over the baseline \cite{bhardwaj2020hostility}. The baseline used the SVM model along with mBERT embeddings. The CNN+BiLSTM models perform better than individual CNN and Bi-LSTM based models. There is a trade-off between coarse-grained F1-scores and the fined-grained scores as an increase in one leads to a decrease in another. The pre-trained embeddings do not show a clear advantage over random initialization. This may be because the target dataset of social media posts is quite different than Wikipedia and News datasets on which these embeddings were trained. Overall BERT based models perform slightly better than basic models.


\section{Conclusion}

In this paper, we evaluate different deep learning approaches for the Constraint @AAAI 2021 Hindi Hostility detection dataset. The task can be seen as a multi-label classification task mainly for multiple categories of hostility type. A combination of different deep learning models along with different variations of word embeddings is compared in this paper. The different models compared in this paper simple 1D CNN, multi-channel CNN, bi-directional LSTM, CNN+LSTM, mBERT, and IndicBERT. The random and fast text embeddings released by IndicNLP and Facebook are the variants of word embeddings used. Through our experiments, we show that basic models perform competitively with BERT based models although the latter is slightly better.  The IndicBERT and mBERT perform equally well. The multi-CNN model when combined with IndicNLP FastText word embedding performs best among the basic models.  

\section*{Acknowledgements} This work was done under the L3Cube Pune mentorship program. We would like to express our gratitude towards our mentors at L3Cube for their continuous support and encouragement.

\bibliographystyle{IEEEtran}
\bibliography{main}

\end{document}